\documentclass[10pt,twocolumn,letterpaper]{article}

\usepackage{cvpr}              

\usepackage{tikz}
\usepackage{pgfplots}
\pgfplotsset{compat=1.18}
\usepgfplotslibrary{fillbetween}
\usetikzlibrary{arrows.meta, positioning, shapes.geometric, fit, calc}
\usepackage{multirow}
\usepackage{colortbl}
\usepackage{tikz}
\usepackage{xcolor}
\usetikzlibrary{arrows.meta,positioning,calc,shadows,backgrounds,fit}

\definecolor{cvprblue}{rgb}{0.21,0.49,0.74}
\usepackage[pagebackref,breaklinks,colorlinks,allcolors=cvprblue]{hyperref}

\def\paperID{18} 
\def\confName{CVPR}
\def\confYear{2026}

\title{AI-Assisted Competency Assessment from Egocentric Video in Simulation-Based Nursing Education}

\author{
Hanchen David Wang$^{*}$\quad
Yilin Liu\thanks{Equal contribution.}\quad
Madison J. Lee\quad
Surya Chand Rayala\\
Gautam Biswas\quad
Daniel T. Levin\quad
Meiyi Ma\\
Vanderbilt University\\
{\tt\small \{hanchen.wang.1, yilin.liu.1, madison.j.lee,}\\
{\tt\small surya.chand.rayala, gautam.biswas, daniel.t.levin, meiyi.ma\}@vanderbilt.edu}
}
\begin{document}
\maketitle

\begin{abstract}
Assessing learner competency in clinical simulation requires expert observation that is time-intensive, difficult to scale, and subject to inter-rater variability. Vision-language models have emerged as a promising tool for understanding complex visual behavior. In this work, we investigate whether visual observations can provide educationally meaningful signals for competency assessment through a three-stage framework that (1)~extracts action timelines from egocentric nursing simulation video using frozen visual encoders and few-shot learning, (2)~derives sequence-level features and per-session recognition metrics, and (3)~relates these to instructor-rated competency. Across 22 densely annotated sessions (3.8 hours, 493 actions), a frozen DINOv2 backbone with HMM Viterbi decoding achieves 57.4\% MOF in leave-one-out 1-shot recognition. Surprisingly, we observe a negative trend between recognition accuracy and competency ($\rho = -0.524$, $p = 0.012$ for mIoU), robust to six confound controls: more competent students produce diverse, harder-to-classify workflows, while simple sequence features show no such relationship. Per-item analysis identifies patient safety protocols and team communication as the expected behaviors most reflected in this pattern, and process model comparisons reveal that higher-competency students exhibit more protocol-consistent action transitions. These findings suggest that recognition accuracy may complement predicted action timelines as a pedagogically informative signal in automated competency assessment.
\end{abstract}

\section{Introduction}
\label{sec:intro}

Across education and workforce training, a central goal is to determine whether learners have developed the knowledge, skills, and judgment needed to perform effectively in practice, a quality broadly termed \emph{competency}~\cite{pellegrino2001knowing}. In domains defined by skilled physical performance, competency assessment requires expert observation of context-dependent behaviors that unfold over time. The consequences of undetected gaps are especially acute in clinical education, where medication administration errors remain among the most common preventable adverse events, often rooted in procedural lapses missed during training~\cite{hayden2014ncsbn}. Simulation-based learning addresses this by letting students practice clinical skills without risk to real patients~\cite{jeffries2005framework}, but competency encompasses not just executing procedures correctly but doing so in an appropriate sequence, with complete safety checks and fluid task transitions~\cite{ericsson1993role}. Instructors assess each session using standardized instruments such as the Creighton Competency Evaluation Instrument (C-CEI)~\cite{todd2008ccei} that map observable behaviors to competency constructs (e.g., clinical judgment, patient safety)~\cite{lasater2007clinical}. This model faces two structural constraints: expert observation cannot scale with growing cohorts~\cite{booth2023engagement}, and inter-rater reliability remains only moderate to substantial even among trained faculty~\cite{husebo2013videobased}. These limitations motivate automated approaches for competency assessment that can analyze observable behavior consistently and at scale. In nursing simulation, video provides a rich record of learner performance, capturing procedural actions and their temporal organization. In this work, we investigate the extent to which visual evidence in simulation videos can support competency assessment.

\begin{figure*}[t]
\centering
\includegraphics[width=\linewidth]{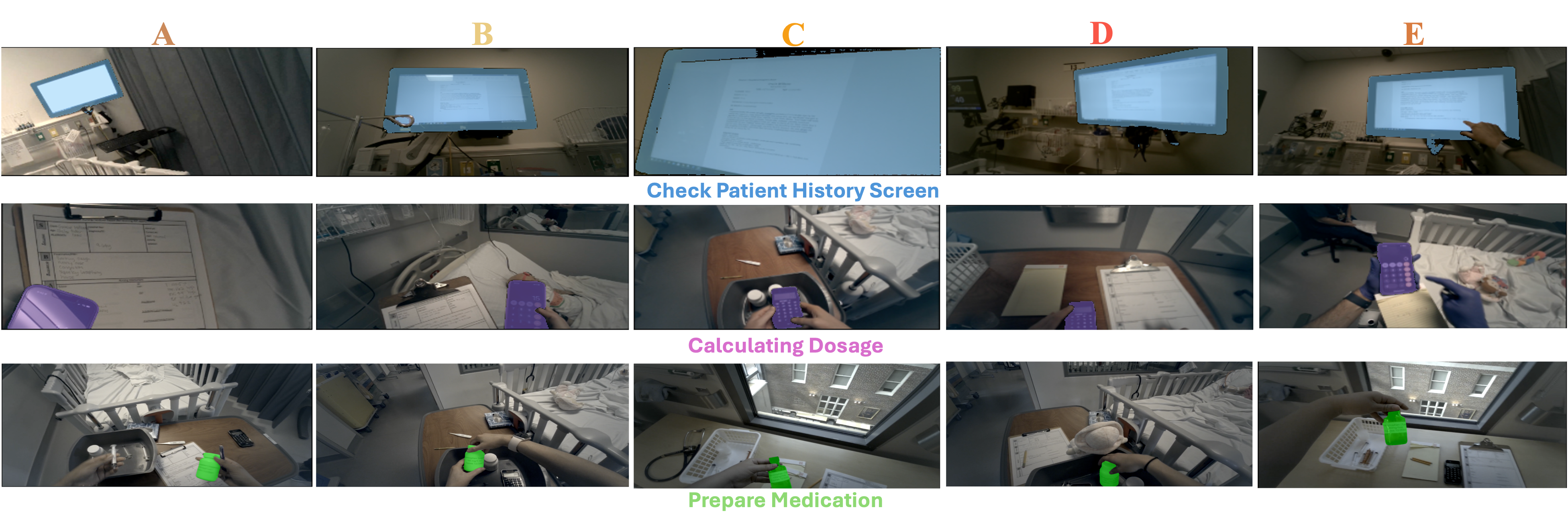}
\caption{Example images of \textcolor{blue!70}{checking the patient screen}, \textcolor{purple!70}{calculating dosage}, and \textcolor{green!70}{preparing medication} from simulation videos of five nursing students. Students A, B, and E use phones for dosage calculation, whereas students C and D use handheld calculators. During medication preparation, students C and E use a dark brown medicine bottle, while students A, B, and D each use a different bottle and hold it differently. More details of the simulation procedure are in App.~\ref{app:scenario}.}
\label{fig:motivating}
\end{figure*}

Video captures what learners do, how they move, and what objects they interact with, all without requiring instrumented environments. Recent advances in first-person video understanding have made egocentric recordings especially compelling. Head-mounted cameras provide an unobstructed, hands-proximal view of what a student attends to and acts upon, and when paired with gaze sensing~\cite{hutt2019automated}, can reveal attentional patterns linked to errors in skilled activities~\cite{wang2023physiq}. As shown in Fig.~\ref{fig:motivating}, this perspective captures the full behaviors of a clinical encounter, from how students hold a medication bottle to which device they use for dosage calculation, preserving precisely the signatures that distinguish novice from expert workflows.

Medication administration is an ideal proving ground because competence is inherently sequential: correct actions in the wrong order, or with safety steps omitted, constitute a clinical error~\cite{jeffries2005framework}. The annotation codebook used in this study was developed around the medication administration workflow, drawing on established nursing competency measurement tools~\cite{schroers2025tool}, and captures fine-grained actions such as dosage calculation. The C-CEI rubric maps expected behaviors to broader competency constructs across the full clinical encounter; vision-based analysis can speak directly to the former, while the latter requires instructor judgment. Applying egocentric video understanding to this setting, however, introduces domain-specific challenges: clinical action vocabularies are absent from standard benchmarks, cohorts are small due to privacy constraints, and models pre-trained on real bodies face a domain gap with simulation mannequins. While surgical education has demonstrated that AI can recognize operative phases and classify skill levels from laparoscopic recordings~\cite{khalifa2025automated}, that work assumes data-rich, fixed-camera environments. Nursing simulation differs in that the recording is egocentric, cohorts are governed by IRB constraints, and competency is holistic rather than tied to a single technical procedure. Yet these difficulties may not be purely noise. Research in surgical skill assessment has found that automated classifiers perform worse on higher-skilled practitioners~\cite{soleymani2021surgical}, and that temporal patterns of action execution capture skill level better than outcome measures alone~\cite{yen2025cholec}. This raises the possibility that recognition accuracy itself carries a pedagogical signal, with lower accuracy reflecting the diverse workflows of more competent students.

This gap motivates the present study, which investigates whether egocentric video, analyzed through frozen visual encoders and few-shot learning, can support competency assessment in nursing simulation. For such an approach to be useful in simulation education, it must capture observable learner performance, reflect differences that are meaningful for instructor-rated competency, and reveal how task execution differs across learners. Accordingly, we investigate the following three \textbf{research questions} in this study:

\begin{enumerate}[label=\textbf{RQ\arabic*.},leftmargin=*,nosep]
    \item To what extent can few-shot action recognition identify clinical actions in egocentric nursing simulation video?
    \item To what extent do automatically extracted action sequences and recognition difficulty relate to instructor-rated competency, and which expected behaviors on the C-CEI are most reflected in vision-based action analysis?
    \item What temporal action patterns distinguish higher- and lower-performing students?
\end{enumerate}

\noindent\textbf{Contributions.} To answer these questions, we propose a three-stage framework that extracts action timelines, analyzes their sequential structure, and relates sequence features and recognition accuracy to competency scores across 22 densely annotated sessions. Our specific contributions are as follows:
\begin{enumerate}[leftmargin=*,nosep]
    \item \textbf{Few-shot clinical action recognition.} We show that frozen DINOv2 features with HMM Viterbi decoding achieve 57.4\% MOF in leave-one-out 1-shot action recognition of egocentric nursing simulation video, establishing feasibility under extremely low-data conditions without any fine-tuning.
    \item \textbf{Classification difficulty to competency.} We observe a negative trend between recognition accuracy and instructor-rated competency ($\rho = -0.524$, $p = 0.012$ for mIoU), robust to six confound controls. Per-item analysis identifies expected behaviors related to patient safety protocols and team communication as 
    the C-CEI items most reflected in this pattern.
    \item \textbf{Temporal workflow analysis.} Process model comparisons from ground-truth action sequences show that higher-performing students exhibit more diverse, protocol-consistent action transitions, delineating the boundaries of unimodal video-based assessment.
\end{enumerate} 

\section{Related Work}
\label{sec:related}

\noindent\textbf{Computer vision for clinical skill assessment.}
Deep learning has been applied extensively to surgical workflow recognition and skill evaluation from operative video, including tool detection and phase recognition~\cite{liu2025smartseg}, direct skill classification from video~\cite{funke2019video}, fine-grained action triplet recognition, and scalable objective assessment of technical skill~\cite{hashimoto2018cv}. These studies highlight the promise of video-based clinical assessment, but most are developed in data-rich surgical settings with fixed cameras and relatively controlled workflows. In contrast, our setting is egocentric and small-scale, and the goal is to assess holistic nursing competency rather than isolated technical skill.
\noindent\textbf{Computer vision for education and learning analytics.}
MMLA integrates video, audio, physiological signals, and interaction logs to study learning processes. Within this paradigm, vision has been used to detect learning-relevant affective states~\cite{bosch2015automatic}, align neural attention with human gaze~\cite{sood2023multimodal}, analyze embodied classroom learning~\cite{ fonteles2026analyzing}, and model student interaction sequences; a recent review surveys multimodal methods across adult training environments, including nursing simulation~\cite{cohn2024multimodal}. We focus on a single modality (egocentric video) to establish what vision alone can reveal about clinical competency. \noindent\textbf{Few-shot and temporal action recognition.}
Prototype networks enable classification with minimal labeled examples. Temporal action segmentation has advanced rapidly on standard benchmarks for different modalities, and large-scale egocentric datasets together with self-supervised encoders such as DINOv2~\cite{oquab2024dinov2} provide strong frozen representations. We combine prototype matching with HMM Viterbi decoding~\cite{rabiner1989tutorial} for temporal segmentation under extremely low-data clinical conditions.

\section{Problem Formulation}

\begin{figure*}[!t]
\centering
\begin{tikzpicture}[>=Stealth, node distance=0.35cm,
    stage/.style={draw, rounded corners=3pt, minimum height=0.6cm, minimum width=1.2cm, font=\scriptsize, align=center, thick},
    data/.style={stage, fill=blue!8},
    proc/.style={stage, fill=orange!12},
    outn/.style={stage, fill=green!10},
    ext/.style={stage, fill=gray!15},
    lbl/.style={font=\scriptsize\bfseries, text=gray!70!black},
    arr/.style={->, thick, color=gray!60!black},
    darr/.style={->, thick, dashed, color=gray!50!black},
    stagebox/.style={draw, dashed, rounded corners=5pt, gray!60, thick, inner sep=6pt},
]
\node[ext] (video) {Egocentric\\Video $V_i$};

\node[proc, right=0.7cm of video, minimum width=2.2cm] (model) {Few-Shot Temporal\\Action Modeling};
\node[outn, right=of model] (preds) {Predicted\\Timeline $\hat{\mathbf{y}}_i$};
\node[data, above=0.3cm of preds] (gt) {Ground-Truth\\$\mathbf{y}_i^*$};

\draw[arr] (video) -- (model);
\draw[arr] (model) -- (preds);
\draw[darr] (gt) -- (preds);

\node[stagebox, fit=(model)(preds)(gt), label={[lbl, anchor=south west]north west:Stage 1: Action Recognition}] (box1) {};

\node[proc, right=0.7cm of preds] (seg) {Segment\\Extraction};
\node[outn, right=of seg] (feat) {Action\\Sequence $\mathbf{s}_i$};
\coordinate (s2top) at (feat.north |- gt.north);

\draw[arr] (preds) -- (seg);
\draw[arr] (seg) -- (feat);

\draw[darr] (gt.east) -- (gt.east -| feat) -- (feat.north);

\node[stagebox, fit=(seg)(feat)(s2top), label={[lbl, anchor=south west]north west:Stage 2: Sequence Analysis}] (box2) {};

\node[proc, right=0.7cm of feat] (corr) {Pattern\\Analysis};
\node[outn, right=of corr] (comp) {Competency\\Assessment};
\node[data, above=0.3cm of comp] (instr) {Instructor\\Scores $\mathbf{c}_i$};
\coordinate (s3top) at (instr.north |- gt.north);

\draw[arr] (feat) -- (corr);
\draw[arr] (corr) -- (comp);
\draw[darr] (instr) -- (comp);

\node[stagebox, fit=(corr)(comp)(instr)(s3top), label={[lbl, anchor=south west]north west:Stage 3: Competency Analysis}] (box3) {};

\end{tikzpicture}
\caption{Overview of the proposed three-stage framework. Gray boxes denote inputs, orange boxes denote processing modules, green boxes denote outputs, and blue boxes denote supervision or reference signals. Solid arrows indicate the forward inference flow, while dashed arrows indicate supervision or oracle guidance. The three stages perform action timeline prediction, action sequence construction, and competency assessment, respectively.} 
\label{fig:pipeline}
\end{figure*}
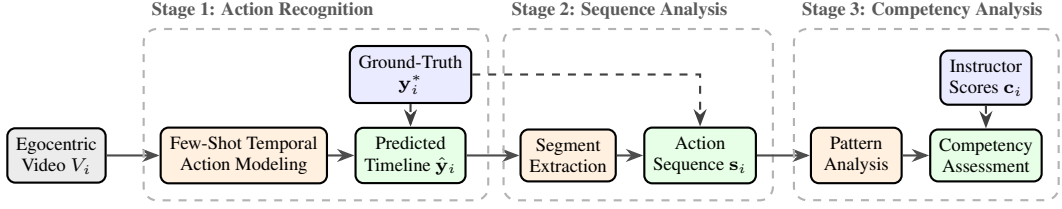

\label{sec:formulation}

Let $\mathcal{V} = \{V_1, \ldots, V_N\}$ denote a set of $N$ egocentric video sessions, where each session $V_i = (f_i^{(1)}, f_i^{(2)}, \ldots, f_i^{(T_i)})$ consists of $T_i$ ordered frames, where $i \in \{1,\ldots,N\}$ indexes the session and $t \in \{1,\ldots,T_i\}$ indexes the frame. Each video is associated with an instructor-assigned competency score vector $\mathbf{c}_i \in \mathbb{R}^{23}$ across 23 expected behaviors on the C-CEI rubric (App.~\ref{sec:appendix_rubric}), of which 11 correspond to video-observable actions; the mean of these 11 items yields the overall competency percentage used for association analyses. Not all items are rated for every session, so some entries of $\mathbf{c}_i$ are missing.

\noindent\textit{Stage 1: Action Recognition.}
Given a clinically grounded action taxonomy $\mathcal{A} = \{a_1, \ldots, a_K, a_\varnothing\}$ comprising $K{=}16$ clinical action classes and one background class $a_\varnothing$ (17 labels total), the goal is to assign each frame a label  $y_i^{(t)} \in \mathcal{A}$, producing a frame-level prediction $\hat{\mathbf{y}}_i = (\hat{y}_i^{(1)}, \ldots, \hat{y}_i^{(t)})$. A frozen encoder $\phi$ extracts per-frame features $\mathbf{z}_i^{(t)} = \phi\big(f_i^{(t)}\big)$, which are matched against class prototypes computed from a support set $\mathcal{S}$ of labeled exemplars sampled from held-out sessions:
\begin{equation}
    \hat{\mathbf{y}}_i = \text{Decode}\bigl(\text{sim}(\phi(V_i),\; \mathcal{P}(\mathcal{S}))\bigr),
    \label{eq:fewshot}
\end{equation}
where $\mathcal{P}(\mathcal{S})$ computes class prototypes from the support set~\cite{snell2017prototypical} (Sec.~\ref{sec:method_stage1}) and $\text{Decode}$ applies HMM Viterbi decoding~\cite{rabiner1989tutorial} to enforce temporally coherent label sequences.
We evaluate $\hat{\mathbf{y}}_i$ against ground-truth annotations $\mathbf{y}_i^*$ using frame-level accuracy (MOF), mean intersection-over-union (mIoU), and segmental F1 (RQ1).

\noindent\textit{Stage 2: Sequence Analysis.} From the predicted frame-level labels $\hat{\mathbf{y}}_i$, we collapse 
contiguous same-label frames into an ordered action sequence 
$\mathbf{s}_i = \bigl((c_i^{(1)}, d_i^{(1)}), \ldots, (c_i^{(L_i)}, d_i^{(L_i)})\bigr)$ 
of $L_i$ segments ($l \in \{1,\ldots,L_i\}$), where 
$c_i^{(l)} \in \mathcal{A} \setminus \{a_\varnothing\}$ is the action label 
and $d_i^{(l)}$ is the segment duration in frames.
From this sequence we derive two families of features used in subsequent analyses: (1)~action transition frequencies, which capture the pairwise flow between clinical actions, and (2)~per-video recognition metrics (MOF, mIoU, F1), which summarize how well the classifier fits each session.

\noindent\textit{Stage 3: Competency Analysis.}
Given the small sample size ($N = 22$) and the pedagogical requirement for transparent feedback, we map sequence features to competency scores using Spearman rank association. To disentangle action detection errors from the intrinsic limits of vision-based assessment, we evaluate under both oracle (features from ground-truth $\mathbf{y}_i^*$) and predicted (features from $\hat{\mathbf{y}}_i$) conditions. Per-item analysis examines which expected behaviors are captured by action sequences alone, and comparison of ground-truth action transition graphs across performance groups identifies discriminative temporal patterns (RQ2, RQ3).

\section{Method}
\label{sec:method}


We collect egocentric video from 22 nursing students, each performing a single standardized pediatric simulation on high-fidelity mannequins. Each session captures one student's complete first-person view of the clinical encounter, recorded via egocentric glasses at 25 FPS. Sessions range from 4 to 24 minutes (mean 10.5 min, total 3.8 hours). All human subjects' data were collected under IRB-approved protocols (\#211801) with informed consent. 

Each session is annotated across three temporal layers by a trained coder using the NOVA annotation system: (1) Behaviors (3 classes: Introduction, Assessment, Administration), (2) Actions ($K = 16$ fine-grained clinical classes plus one background class $a_\varnothing$ for unannotated frames, yielding 17 labels total; see App.~\ref{sec:appendix_actions}), and (3) Communication (Patient, Family, Provider). The Action layer, containing 493 annotated clinical segments, serves as the primary target for few-shot recognition. Each session is independently rated by an expert instructor across 23 expected behaviors using the C-CEI rubric (App.~\ref{sec:appendix_rubric}), of which the 11 video-observable items yield the overall competency percentage used throughout this study. Inter-rater reliability was assessed on 3 stratified videos (low/median/high competency) independently annotated by a second rater, yielding substantial agreement (mean Cohen's $\kappa = 0.708$; App.~\ref{app:irr}).

\subsection{Stage 1: Action Recognition}
\label{sec:method_stage1}

\noindent\textbf{Feature extraction.}
We extract frame-level features from each video using a frozen backbone encoder. For each frame $f_i^{(t)}$, we obtain a feature vector $\mathbf{z}_i^{(t)} = \phi\big(f_i^{(t)}\big) \in \mathbb{R}^D$, followed by L2 normalization: $\mathbf{z}_i^{(t)} \leftarrow \mathbf{z}_i^{(t)} / \|\mathbf{z}_i^{(t)}\|$. We evaluate three backbones: (1) ResNet-50~\cite{he2016deep} (ImageNet-supervised, $D = 2048$), (2) DINOv2 ViT-B/14~\cite{oquab2024dinov2} (ImageNet self-supervised, $D = 768$), and (3) CLIP ViT-B/16~\cite{radford2021learning} (vision-language contrastive, $D = 512$). All backbones are frozen with no fine-tuning.

\noindent\textbf{Prototype computation.}
In the cross-sample (leave-one-out) setting, we construct class prototypes from the $N{-}1$ support sessions following the prototypical network paradigm. For each support session $V_j$ and each class $k$ present in that session, we randomly sample $n$ labeled frames and compute a per-session centroid $\boldsymbol{\mu}_{k,j}$, which is then L2-normalized. Because not every session contains every action class, the global prototype for class $k$ is obtained by averaging the normalized centroids only over sessions that contain that class: $\mathbf{p}_k = \frac{1}{|\mathcal{J}_k|}\sum_{j \in \mathcal{J}_k} \frac{\boldsymbol{\mu}_{k,j}}{\|\boldsymbol{\mu}_{k,j}\|}$, where $\mathcal{J}_k$ is the set of sessions containing class $k$. The per-session normalization ensures each session contributes a unit-direction vector, preventing sessions whose sampled frames are more self-consistent from dominating the prototype direction. The aggregated prototype is then L2-normalized again to ensure it lies on the unit sphere, since the mean of unit vectors is not itself unit-length in general. As an alternative, we also evaluate a clustered strategy in which all support frames for each class are pooled and partitioned into $k{=}3$ sub-centroids via $k$-means; each query frame is then assigned to the class of its nearest sub-centroid.

\noindent\textbf{Classification.}
Each query frame is scored against all prototypes (including the background class $a_\varnothing$) via cosine similarity. Rather than committing to a hard per-frame label at this stage, the continuous similarity scores are passed directly to the temporal smoothing step below, which jointly optimizes over the entire sequence.

\noindent\textbf{Temporal smoothing.}
We apply \emph{HMM Viterbi decoding}~\cite{rabiner1989tutorial} to enforce temporally coherent label sequences. The transition matrix $\mathbf{A}$ is learned from support session label sequences with Laplace smoothing, prior probabilities $\boldsymbol{\pi}$ are estimated from action start frequencies, and emission log-probabilities are computed as temperature-scaled ($\tau = 5$, selected via grid search on held-out folds; values in the range 5--10 are standard for cosine-prototype matching in few-shot settings~\cite{snell2017prototypical}) log-softmax of cosine similarities:

\begin{equation}
\begin{aligned}
\log P(\mathbf{z}_i^{(t)} \mid a_k) 
&= \tau \cdot \text{cos}\big(\mathbf{z}_i^{(t)}, \mathbf{p}_k\big) \\
&\quad - \log \sum_{k'} \exp\!\Big(\tau \cdot \text{cos}\big(\mathbf{z}_i^{(t)}, \mathbf{p}_{k'}\big)\Big).
\end{aligned}
\end{equation}
The Viterbi algorithm~\cite{rabiner1989tutorial} then recovers the optimal label sequence $\hat{\mathbf{y}}_i = (\hat{y}_i^{(1)}, \ldots, \hat{y}_i^{(T_i)})$ by selecting, at each time step, the action label that jointly maximizes the cumulative sum of emission log-probabilities and transition log-probabilities over the entire sequence, thereby enforcing clinically plausible transitions rather than treating each frame independently. This smoothed timeline is the final output of Stage~1 (Sec.~\ref{sec:method_stage1}) and the prediction used in all subsequent analyses (Sec.~\ref{sec:method_stage2}--\ref{sec:method_stage3}).

\subsection{Stage 2: Sequence Analysis}
\label{sec:method_stage2}

From the frame-level predictions $\hat{\mathbf{y}}_i$, we collapse contiguous same-label frames into an ordered sequence of action segments via run-length encoding, discarding segments below a minimum duration threshold and removing background segments. From the resulting clinical action sequence we compute two families of features. First, we tabulate pairwise action transition frequencies, which record how often each action class is followed by every other class within a session; these frequencies form the basis of the process model comparison in Sec.~\ref{sec:results_rq3}. Second, we retain the per-video frame-level recognition metrics (MOF, mIoU, F1) computed during Stage~1, which serve as summary measures of how well the classifier fits each session and are used in the competency analysis (Sec.~\ref{sec:results_rq2}).

\subsection{Stage 3: Competency Analysis}
\label{sec:method_stage3}

Given the small sample size and the pedagogical need for transparency, we employ Spearman's rank correlation to test the relationship between sequence-level features and video-observable competency scores (11 items). To assess robustness, we compute partial associations controlling for potential confounds (annotation coverage, video duration, segment count). Per-item analysis examines which of the 23 expected behaviors are most reflected in recognition accuracy. Process model analysis (Heuristics Miner) visualizes action transition graphs for higher- and lower-competency groups to identify differences in clinical workflows.
\section{Evaluation Results}
\label{sec:results}

\subsection{RQ1: Few-Shot Clinical Action Recognition}
\label{sec:results_rq1}

We evaluate few-shot action recognition under two settings: \emph{within-sample}, where support and query frames are drawn from the same video, and \emph{cross-sample} (leave-one-out), where the model must generalize to entirely unseen sessions. We report three standard temporal action segmentation metrics: mean-over-frames accuracy (MOF), mean intersection-over-union (mIoU), and segmental F1 score.

\subsubsection{Within-Sample Evaluation}

In the within-sample setting, for each of the 22 videos, $n$ frames per action class are randomly sampled as support prototypes (where $n$ denotes the shot count); the remaining frames serve as the query set. Tab.~\ref{tab:within_sample} reports within-sample performance across five shot counts. Recognition quality improves substantially with more support examples, with the largest gain between 1 and 3 shots. Performance plateaus around 10--15 shots, indicating that even a modest number of labeled exemplars enables reliable within-sample segmentation and that the bottleneck lies in cross-session generalization rather than representation capacity.

\begin{table*}[ht]
\centering
\caption{Within-sample few-shot action recognition. For each video, $n$ frames per action class are sampled as prototypes, and the remaining frames serve as the query set. Results are reported as mean $\pm$ std over 22 videos. For all backbones, \textbf{bold} indicates the best-performing configuration per metric and \underline{underlined} indicates the second-best. Higher is better for all metrics.}
\label{tab:within_sample}
\scriptsize
\setlength{\tabcolsep}{3pt}
\renewcommand{\arraystretch}{1.05}
\begin{tabular}{lccccccccc}
\toprule
\multirow{2}{*}{\textbf{Shots ($n$)}}
& \multicolumn{3}{c}{\textbf{DINOv2}}
& \multicolumn{3}{c}{\textbf{ResNet50}}
& \multicolumn{3}{c}{\textbf{CLIP (ViT-B/16)}} \\
\cmidrule(r){2-4} \cmidrule(lr){5-7} \cmidrule(l){8-10}
& \textbf{MOF} & \textbf{mIoU} & \textbf{F1}
& \textbf{MOF} & \textbf{mIoU} & \textbf{F1}
& \textbf{MOF} & \textbf{mIoU} & \textbf{F1} \\
\midrule
1
& 0.555 $\pm$ 0.125 & 0.419 $\pm$ 0.108 & 0.522 $\pm$ 0.115
& 0.559 $\pm$ 0.150 & 0.416 $\pm$ 0.130 & 0.514 $\pm$ 0.140
& 0.556 $\pm$ 0.135 & 0.393 $\pm$ 0.113 & 0.495 $\pm$ 0.124 \\
3
& 0.783 $\pm$ 0.083 & 0.632 $\pm$ 0.088 & 0.736 $\pm$ 0.077
& 0.786 $\pm$ 0.075 & 0.637 $\pm$ 0.078 & 0.737 $\pm$ 0.072
& 0.797 $\pm$ 0.081 & 0.642 $\pm$ 0.086 & 0.744 $\pm$ 0.076 \\
5
& 0.847 $\pm$ 0.072 & 0.715 $\pm$ 0.099 & 0.805 $\pm$ 0.076
& 0.836 $\pm$ 0.079 & 0.699 $\pm$ 0.106 & 0.791 $\pm$ 0.084
& 0.828 $\pm$ 0.082 & 0.685 $\pm$ 0.104 & 0.779 $\pm$ 0.086 \\
10
& \underline{0.905 $\pm$ 0.054} & \underline{0.797 $\pm$ 0.081} & \underline{0.852 $\pm$ 0.067}
& \underline{0.896 $\pm$ 0.056} & \underline{0.795 $\pm$ 0.074} & \underline{0.852 $\pm$ 0.065}
& \underline{0.908 $\pm$ 0.041} & \underline{0.798 $\pm$ 0.069} & \underline{0.850 $\pm$ 0.067} \\
15
& \textbf{0.930 $\pm$ 0.042} & \textbf{0.846 $\pm$ 0.077} & \textbf{0.869 $\pm$ 0.078}
& \textbf{0.923 $\pm$ 0.044} & \textbf{0.833 $\pm$ 0.090} & \textbf{0.856 $\pm$ 0.090}
& \textbf{0.931 $\pm$ 0.030} & \textbf{0.844 $\pm$ 0.066} & \textbf{0.858 $\pm$ 0.064} \\
\bottomrule
\end{tabular}
\end{table*}

\subsubsection{Cross-Sample Evaluation}

The more challenging and practically relevant setting is cross-sample evaluation, where the model must generalize to entirely unseen sessions with unseen participants. We adopt a leave-one-out protocol across all 22 sessions: in each of 22 folds, one session is held out as the query video, and class prototypes are constructed from the remaining 21 support sessions using prototype computation (Sec.~\ref{sec:method_stage1}). We vary the shot count $n \in \{1,3,5,10,15\}$ frames sampled per class per session to examine how prototype quality scales with the support budget. The query video is classified frame-by-frame via cosine similarity followed by HMM Viterbi decoding.

\begin{table*}[ht]
\centering
\caption{Cross-sample few-shot action recognition (leave-one-out, 22 folds). For each held-out video, $n$ frames per action class are sampled from the 21 support sessions to construct prototypes, and the held-out session is classified via HMM Viterbi decoding. Results are reported as mean $\pm$ std over 22 folds. \textbf{Bold} indicates the best-performing configuration per metric within each backbone, and \underline{underlined} indicates the second-best. Higher is better for all metrics.}
\label{tab:backbone_comparison_final}
\scriptsize
\setlength{\tabcolsep}{3pt}
\renewcommand{\arraystretch}{1.05}
\begin{tabular}{llccccccccc}
\toprule
\multirow{2}{*}{\textbf{Shots ($n$)}} & \multirow{2}{*}{\textbf{Proto.}}
& \multicolumn{3}{c}{\textbf{DINOv2}}
& \multicolumn{3}{c}{\textbf{ResNet50}}
& \multicolumn{3}{c}{\textbf{CLIP (ViT-B/16)}} \\
\cmidrule(r){3-5} \cmidrule(lr){6-8} \cmidrule(l){9-11}
& & \textbf{MOF} & \textbf{mIoU} & \textbf{F1}
& \textbf{MOF} & \textbf{mIoU} & \textbf{F1}
& \textbf{MOF} & \textbf{mIoU} & \textbf{F1} \\
\midrule

\multirow{2}{*}{1}
& Mean
& 0.574 $\pm$ 0.121 & 0.337 $\pm$ 0.088 & 0.341 $\pm$ 0.092
& 0.467 $\pm$ 0.139 & 0.255 $\pm$ 0.083 & 0.256 $\pm$ 0.073
& 0.533 $\pm$ 0.134 & 0.306 $\pm$ 0.098 & 0.322 $\pm$ 0.095 \\
& Clust.
& 0.472 $\pm$ 0.111 & 0.280 $\pm$ 0.088 & 0.245 $\pm$ 0.061
& 0.397 $\pm$ 0.126 & 0.225 $\pm$ 0.084 & 0.195 $\pm$ 0.055
& 0.464 $\pm$ 0.160 & 0.292 $\pm$ 0.105 & 0.255 $\pm$ 0.073 \\
\midrule

\multirow{2}{*}{3}
& Mean
& 0.622 $\pm$ 0.128 & 0.433 $\pm$ 0.122 & 0.412 $\pm$ 0.106
& \underline{0.507 $\pm$ 0.113} & \underline{0.302 $\pm$ 0.078} & 0.287 $\pm$ 0.058
& 0.580 $\pm$ 0.137 & 0.405 $\pm$ 0.132 & 0.373 $\pm$ 0.111 \\
& Clust.
& 0.569 $\pm$ 0.104 & 0.403 $\pm$ 0.103 & 0.342 $\pm$ 0.084
& 0.445 $\pm$ 0.106 & 0.257 $\pm$ 0.064 & 0.226 $\pm$ 0.063
& 0.502 $\pm$ 0.112 & 0.337 $\pm$ 0.108 & 0.279 $\pm$ 0.081 \\
\midrule

\multirow{2}{*}{5}
& Mean
& 0.640 $\pm$ 0.119 & 0.436 $\pm$ 0.109 & 0.411 $\pm$ 0.096
& \textbf{0.508 $\pm$ 0.139} & \textbf{0.327 $\pm$ 0.111} & \textbf{0.304 $\pm$ 0.087}
& 0.596 $\pm$ 0.115 & 0.405 $\pm$ 0.099 & 0.390 $\pm$ 0.099 \\
& Clust.
& 0.594 $\pm$ 0.134 & 0.423 $\pm$ 0.115 & 0.372 $\pm$ 0.107
& 0.394 $\pm$ 0.110 & 0.247 $\pm$ 0.086 & 0.224 $\pm$ 0.061
& 0.540 $\pm$ 0.136 & 0.371 $\pm$ 0.098 & 0.333 $\pm$ 0.083 \\
\midrule

\multirow{2}{*}{10}
& Mean
& \textbf{0.656 $\pm$ 0.152} & \textbf{0.451 $\pm$ 0.128} & \textbf{0.419 $\pm$ 0.100}
& 0.473 $\pm$ 0.138 & 0.298 $\pm$ 0.085 & 0.279 $\pm$ 0.076
& \textbf{0.618 $\pm$ 0.124} & \textbf{0.439 $\pm$ 0.098} & \textbf{0.414 $\pm$ 0.104} \\
& Clust.
& 0.612 $\pm$ 0.131 & 0.433 $\pm$ 0.117 & 0.407 $\pm$ 0.112
& 0.428 $\pm$ 0.123 & 0.282 $\pm$ 0.101 & 0.268 $\pm$ 0.089
& 0.574 $\pm$ 0.116 & 0.400 $\pm$ 0.084 & 0.374 $\pm$ 0.076 \\
\midrule

\multirow{2}{*}{15}
& Mean
& \underline{0.644 $\pm$ 0.148} & \underline{0.446 $\pm$ 0.128} & \underline{0.417 $\pm$ 0.106}
& 0.496 $\pm$ 0.163 & 0.299 $\pm$ 0.116 & \underline{0.290 $\pm$ 0.109}
& \underline{0.612 $\pm$ 0.134} & \underline{0.426 $\pm$ 0.097} & \underline{0.407 $\pm$ 0.088} \\
& Clust.
& 0.594 $\pm$ 0.128 & 0.399 $\pm$ 0.106 & 0.391 $\pm$ 0.099
& 0.433 $\pm$ 0.146 & 0.256 $\pm$ 0.090 & 0.249 $\pm$ 0.078
& 0.550 $\pm$ 0.149 & 0.385 $\pm$ 0.096 & 0.364 $\pm$ 0.087 \\
\bottomrule
\end{tabular}
\end{table*}

Tab.~\ref{tab:backbone_comparison_final} reports cross-sample performance for DINOv2, ResNet-50 and CLIP across five shot counts (1, 3, 5, 10, 15), comparing mean and clustered prototype strategies. DINOv2 with mean prototypes consistently outperforms all other configurations, achieving its best performance at 10 shots (65.6\% MOF, 45.1\% mIoU, 41.9\% F1). Mean prototypes substantially outperform clustered prototypes across both backbones, indicating that splitting each class into multiple sub-centroids introduces false matches under the few-shot budget. DINOv2 consistently outperforms ResNet-50 and CLIP across all shot counts and metrics, suggesting that self-supervised vision transformer features offer better discrimination of fine-grained hand-object interactions in clinical settings.

Comparing Tab.~\ref{tab:backbone_comparison_final} with Tab.~\ref{tab:within_sample}, the cross-sample setting shows substantially lower performance than within-sample at the same nominal shot count $n$. Importantly, these are not directly comparable in terms of total support data: within-sample uses $n$ frames per class from a single video, whereas cross-sample pools $n$ frames per class from each of 21 sessions, yielding $20$ times more support frames per class overall. Despite this $21{\times}$ larger support budget, cross-sample performance lags behind, underscoring that the gap is driven by participant-level domain shift (differences in student appearance, camera angle, pace, and workflow ordering across individuals) rather than insufficient support data. Yet the following sections show that this recognition variability is itself pedagogically informative.

\subsection{RQ2: Action Sequences, Competency, and Per-Item Analysis}
\label{sec:results_rq2}

To investigate whether extracted action recognition metrics carry information related to instructor-rated competency, we compute Spearman rank associations between per-video recognition metrics obtained from cross-sample (leave-one-out) evaluation using the best model (DINOv2 + HMM, 10-shot) and the overall competency score (mean of the 11 video-observable rubric items; see Sec.~\ref{sec:formulation}).

\subsubsection{Overall Trends}

Tab.~\ref{tab:competency_association} presents a notable pattern: all three recognition accuracy metrics show a negative trend as video-observable competency increases. The strongest observed relationship is for mIoU ($\rho = -0.524$, $p = 0.012$), which measures per-class balance. MOF ($\rho = -0.439$, $p = 0.041$) and F1 ($\rho = -0.433$, $p = 0.044$) show similar patterns. Neither the number of ground-truth action classes nor the number of labeled query frames shows any significant relationship with competency,  ruling out annotation-count artifacts as a confounding explanation.

\begin{table}[ht]
\centering
\caption{Spearman $\rho$ between per-video recognition metrics and overall video-observable competency score (11 items, $N=22$). All accuracy metrics show a negative trend with competency.}
\label{tab:competency_association}
\resizebox{\columnwidth}{!}{%
\begin{tabular}{lccc}
\toprule
Feature & Spearman $\rho$ & $p$-value & Pearson $r$ \\
\midrule
\textbf{mIoU (per-class accuracy)} & $\mathbf{-0.524}$ & $\mathbf{0.012}$ & $-0.469$ \\
\textbf{MOF (frame accuracy)}      & $\mathbf{-0.439}$ & $\mathbf{0.041}$ & $-0.436$ \\
\textbf{F1 (macro)}                & $\mathbf{-0.433}$ & $\mathbf{0.044}$ & $-0.407$ \\
Frame error rate (1$-$MOF)         & $+0.439$ & $0.041$ & $+0.436$ \\
\# Action classes in GT            & $-0.087$ & $0.701$ & $-0.043$ \\
\# Labeled query frames            & $-0.071$ & $0.754$ & $+0.016$ \\
\bottomrule
\end{tabular}%
}
\end{table}

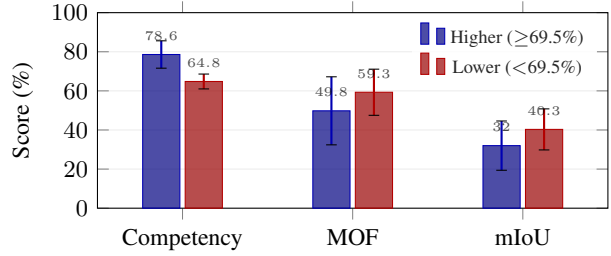
\begin{figure}[t]
\centering
\begin{tikzpicture}
\begin{axis}[
    width=\columnwidth,
    height=0.5\columnwidth,
    ybar=2pt,
    bar width=14pt,
    ymin=0, ymax=100,
    ylabel={Score (\%)},
    symbolic x coords={Competency, MOF, mIoU},
    xtick=data,
    tick label style={font=\small},
    every axis label/.style={font=\small},
    legend style={at={(0.98,0.98)}, anchor=north east, font=\scriptsize, draw=none, fill=white, fill opacity=0.85},
    nodes near coords,
    every node near coord/.append style={font=\tiny, yshift=2pt},
    grid=major,
    grid style={gray!15},
    ymajorgrids=true,
    xmajorgrids=false,
    enlarge x limits=0.25,
]
\addplot[fill=blue!50!black, fill opacity=0.7, draw=blue!70!black,
    error bars/.cd, y dir=both, y explicit, error bar style={thick}]
    coordinates {
        (Competency, 78.6) +- (0, 7.0)
        (MOF, 49.8) +- (0, 17.4)
        (mIoU, 32.0) +- (0, 12.6)
    };
\addlegendentry{Higher ($\geq$69.5\%)}
\addplot[fill=red!60!black, fill opacity=0.7, draw=red!70!black,
    error bars/.cd, y dir=both, y explicit, error bar style={thick}]
    coordinates {
        (Competency, 64.8) +- (0, 3.8)
        (MOF, 59.3) +- (0, 11.8)
        (mIoU, 40.3) +- (0, 10.5)
    };
\addlegendentry{Lower ($<$69.5\%)}
\end{axis}
\end{tikzpicture}
\caption{Group-level comparison: sessions split by median video-observable competency score (11 items). Despite higher instructor ratings, higher-performing students exhibit \emph{lower} classification accuracy. Error bars show $\pm$1 std.}
\label{fig:group_means}
\end{figure}

This pattern is consistent with Moravec's insight: the classifier performs better on the mechanical, templated workflows of lower-performing students, whereas the fluid, adaptive behaviors of higher-performing students are harder to recognize. When sessions are split by median competency score, lower-performing students have 9.5\% higher MOF and 8.3\% higher mIoU than higher-performing students (Fig.~\ref{fig:group_means}). One plausible interpretation, consistent with the motor learning principle of abundance~\cite{ranganathan2020repetition}, is that more competent students perform more diverse workflows with additional safety checks and fluid task transitions, producing greater visual diversity that makes classification harder while earning higher instructor marks. This converges with surgical skill assessment findings where classifiers achieve lower accuracy on higher-skilled practitioners, suggesting that the negative trend in recognition accuracy \emph{may} carry a pedagogically informative signal. We note, however, that with $N = 22$ sessions, this interpretation remains exploratory. Importantly, simple sequence features (screen time ratio, transition count, unique action count) extracted from both oracle and predicted timelines show no significant relationship with competency (all $p > 0.10$).

\subsubsection{Per-Item Analysis}
\label{sec:results_per_item}

To identify which facets of clinical competency are most accessible through vision-based action analysis, we examine Spearman associations between per-video MOF and each of the 23 instructor rubric items. Tab.~\ref{tab:per_item} reports the five items with the strongest associations, ordered by magnitude.

\begin{table}[ht]
\centering
\caption{Top-5 Spearman $\rho$ between per-video MOF and individual rubric items, ordered by magnitude. Except for communication, the remaining behaviors are visually observable. $N$ varies across items because instructors may omit ratings when a behavior is not observed or not applicable during a particular session.}
\label{tab:per_item}
\resizebox{\columnwidth}{!}{%
\begin{tabular}{clccc}
\toprule
\# & Rubric Item & $\rho$ & $p$ & $N$ \\
\midrule
4                      & Communicates effectively with team       & $-0.470$ & $\mathbf{0.049}$ & 18 \\
18 & Uses patient identifiers                  & $-0.455$ & $\mathbf{0.033}$ & 22 \\
19 & Utilizes standardized practices           & $-0.377$ & $0.083$ & 22 \\
21 & Manages technology and equipment          & $-0.350$ & $0.110$ & 22 \\
12 & Prioritizes appropriately                & $-0.343$ & $0.118$ & 22 \\
\bottomrule
\end{tabular}%
}
\end{table}

The per-item trends are broadly similar in magnitude across items, consistent with the expectation that with $N = 22$ sessions, individual rubric items lack sufficient power to differentiate statistically from one another. Two items reach nominal significance: Item~18 (``Uses patient identifiers,'' $\rho = -0.455$, $p = 0.033$) and Item~4 (``Communicates effectively with team,'' $\rho = -0.470$, $p = 0.049$). Patient safety protocols may produce the strongest pattern because students who excel in this expected behavior tend to perform additional wristband-checking and verification steps, generating visually diverse frame sequences that are inherently harder to classify.

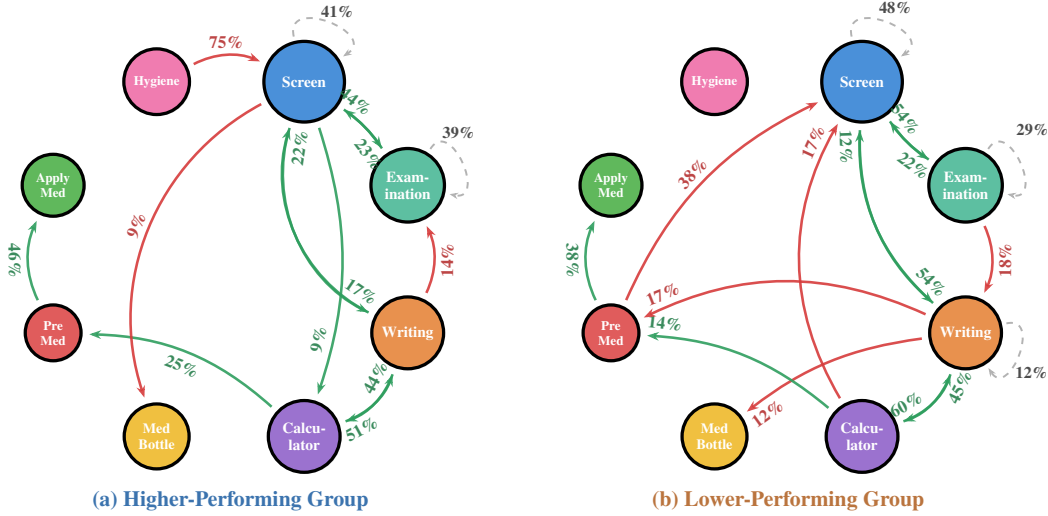
\begin{figure*}[t]
\centering
\scalebox{0.82}{%
\begin{tikzpicture}

\definecolor{nodeScreen}   {HTML}{4A90D9}
\definecolor{nodeExam}     {HTML}{5DBFA1}
\definecolor{nodeWriting}  {HTML}{E8914E}
\definecolor{nodeCalc}     {HTML}{9B72CF}
\definecolor{nodeMedBot}   {HTML}{F0C040}
\definecolor{nodePreMed}   {HTML}{E05C5C}
\definecolor{nodeApplyMed} {HTML}{60B85C}
\definecolor{nodeHygiene}  {HTML}{F07CB0}
\definecolor{arrowGood}    {HTML}{2E9E5E}
\definecolor{arrowBad}     {HTML}{D63B3B}
\definecolor{bgHigh}       {HTML}{EAF3FF}
\definecolor{bgLow}        {HTML}{FFF4EA}
\definecolor{panelBorder}  {HTML}{C8D8E8}

\tikzset{
  mynode/.style={
    circle, draw=black, line width=1.5pt,
    text=white, font=\bfseries\fontsize{7}{8}\selectfont,
    align=center,
    drop shadow={opacity=0.22, shadow xshift=0.7pt, shadow yshift=-0.7pt}
  },
  screen/.style   ={mynode, minimum size=1.30cm, fill=nodeScreen},
  exam/.style     ={mynode, minimum size=1.10cm, fill=nodeExam},
  writing/.style  ={mynode, minimum size=1.10cm, fill=nodeWriting},
  calc/.style     ={mynode, minimum size=1.10cm, fill=nodeCalc},
  bottle/.style   ={mynode, minimum size=0.95cm,
                    font=\bfseries\fontsize{6.5}{7.5}\selectfont, fill=nodeMedBot},
  premed/.style   ={mynode, minimum size=0.90cm,
                    font=\bfseries\fontsize{6}{7}\selectfont, fill=nodePreMed},
  applymed/.style ={mynode, minimum size=0.90cm,
                    font=\bfseries\fontsize{6}{7}\selectfont, fill=nodeApplyMed},
  hygi/.style     ={mynode, minimum size=0.90cm,
                    font=\bfseries\fontsize{6}{7}\selectfont, fill=nodeHygiene},
  ag/.style={-{Stealth[length=5.5pt,width=4pt]},
             line width=1.2pt, draw=arrowGood, opacity=0.9,
             shorten >=3pt, shorten <=3pt},
  ar/.style={-{Stealth[length=5.5pt,width=4pt]},
             line width=1.2pt, draw=arrowBad, opacity=0.9,
             shorten >=3pt, shorten <=3pt},
  labg/.style={font=\bfseries\fontsize{8}{9}\selectfont,
               text=arrowGood!80!black, inner sep=2pt},
  labr/.style={font=\bfseries\fontsize{8}{9}\selectfont,
               text=arrowBad!85!black, inner sep=2pt},
  exitloop/.style={-{Stealth[length=4pt,width=3pt]},
                   line width=0.9pt, draw=gray!60, dashed,
                   shorten >=2pt, shorten <=2pt},
  exitlab/.style={font=\bfseries\fontsize{7.5}{8.5}\selectfont,
                  text=gray!60!black, inner sep=1.5pt},
}

\begin{scope}[xshift=0cm]
  \node[screen]    (S1) at (4.69, 5.46) {Screen};
  \node[exam]      (E1) at (6.36, 3.79) {Exam-\\ination};
  \node[writing]   (W1) at (6.36, 1.41) {Writing};
  \node[calc]      (C1) at (4.69,-0.26) {Calcu-\\lator};
  \node[bottle]    (B1) at (2.31,-0.26) {Med\\Bottle};
  \node[premed]    (P1) at (0.64, 1.41) {Pre\\Med};
  \node[applymed]  (A1) at (0.64, 3.79) {Apply\\Med};
  \node[hygi]      (H1) at (2.31, 5.46) {Hygiene};

  \draw[exitloop] (S1) edge[out=103,in=33,loop,looseness=2.5]
    node[exitlab, pos=0.6, above=3pt] {41\%} (S1);
  \draw[exitloop] (E1) edge[out=58,in=-12,loop,looseness=2.5]
    node[exitlab, pos=0.1, above right=2pt] {39\%} (E1);

  \draw[ag] (P1) to[bend left=25]  node[sloped, below=1pt,  labg, pos=0.5] {46\%} (A1);
  \draw[ag] (C1) to[bend right=20] node[sloped, below=1pt, labg, pos=0.5] {25\%} (P1);
  \draw[ag] (C1) to[bend right=25] node[sloped,below=1pt, labg, pos=0.2] {51\%} (W1);
  \draw[ag] (W1) to[bend left=25]  node[sloped, above=1pt,  labg, pos=0.2] {44\%} (C1);
  \draw[ag] (S1) to[bend left=18] node[sloped, above=1pt,  labg, pos=0.8] {9\%}  (C1);
  \draw[ag] (W1) to[bend left=40]  node[sloped, below=1pt,  labg, pos=0.9] {22\%} (S1);
  \draw[ag] (S1) to[bend right=40] node[sloped, above=1pt, labg, pos=0.9] {17\%} (W1);
  \draw[ag] (S1) to[bend left=10]  node[sloped, above=1pt, labg, pos=0.15] {44\%} (E1);
  \draw[ag] (E1) to[bend right=10]  node[sloped,below=1pt,  labg, pos=0.13] {23\%} (S1);

  \draw[ar] (H1) to[bend left=25]  node[above=1pt, labr, pos=0.5] {75\%} (S1);
  \draw[ar] (S1) to[bend right=40] node[sloped, above=1pt, labr, pos=0.5] {9\%} (B1);
  \draw[ar] (W1) to[bend right=25] node[sloped,below=1pt, labr, pos=0.5] {14\%} (E1);

  \begin{pgfonlayer}{background}
  \end{pgfonlayer}
  \node[font=\bfseries\fontsize{10}{11}\selectfont, text=nodeScreen!80!black]
    at (3.5,-1.3) {(a) Higher-Performing Group};
\end{scope}

\begin{scope}[xshift=9cm]
  \node[screen]    (S2) at (4.69, 5.46) {Screen};
  \node[exam]      (E2) at (6.36, 3.79) {Exam-\\ination};
  \node[writing]   (W2) at (6.36, 1.41) {Writing};
  \node[calc]      (C2) at (4.69,-0.26) {Calcu-\\lator};
  \node[bottle]    (B2) at (2.31,-0.26) {Med\\Bottle};
  \node[premed]    (P2) at (0.64, 1.41) {Pre\\Med};
  \node[applymed]  (A2) at (0.64, 3.79) {Apply\\Med};
  \node[hygi]      (H2) at (2.31, 5.46) {Hygiene};

  \draw[exitloop] (S2) edge[out=103,in=33,loop,looseness=2.5]
    node[exitlab, pos=0.6, above=5pt] {48\%} (S2);
  \draw[exitloop] (E2) edge[out=58,in=-12,loop,looseness=2.5]
    node[exitlab, pos=0.28, above right=4pt] {29\%} (E2);
  \draw[exitloop] (W2) edge[out=13,in=-57,loop,looseness=2.5]
    node[exitlab, pos=0.8, right=5pt] {12\%} (W2);

  \draw[ag] (P2) to[bend left=25]  node[sloped, below=1pt,  labg, pos=0.5] {38\%} (A2);
  \draw[ar] (W2) to[bend right=15] node[sloped, below=1pt, labr, pos=0.9] {12\%} (B2);
  \draw[ag] (C2) to[bend right=25] node[sloped, below=1pt, labg, pos=0.85] {45\%} (W2);
  \draw[ag] (W2) to[bend left=25]  node[sloped, above=1pt,  labg, pos=0.8] {60\%} (C2);
  \draw[ag] (C2) to[bend right=18] node[sloped, above=1pt,  labg, pos=0.9] {14\%} (P2);
  \draw[ag] (S2) to[bend right=25] node[sloped, above=1pt, labg, pos=0.9] {54\%} (W2);
  \draw[ag] (W2) to[bend left=25]  node[sloped, below=1pt,  labg, pos=0.9] {12\%} (S2);
  \draw[ag] (S2) to[bend right=10]  node[sloped, above=1pt, labg, pos=0.2] {54\%} (E2);
  \draw[ag] (E2) to[bend left=10]  node[sloped, below=1pt,  labg, pos=0.2] {22\%} (S2);

  \draw[ar] (C2) to[bend left=30]  node[sloped, above=1pt, labr, pos=0.9] {17\%} (S2);
  \draw[ar] (P2) to[bend left=20]  node[sloped, above=1pt, labr, pos=0.5] {38\%} (S2);
  \draw[ar] (W2) to[bend right=25]  node[sloped, above=1pt, labr, pos=0.9] {17\%} (P2);
  \draw[ar] (E2) to[bend left=25] node[sloped, below=1pt, labr, pos=0.5] {18\%} (W2);

  \begin{pgfonlayer}{background}
  \end{pgfonlayer}
  \node[font=\bfseries\fontsize{10}{11}\selectfont, text=nodeWriting!80!black]
    at (3.5,-1.3) {(b) Lower-Performing Group};
\end{scope}

\end{tikzpicture}
}
\caption{Process models comparing the higher-performing group (video-observable competency score $\geq$69.5\%, the median on 22 sessions) and the lower-performing group ($<$69.5\%) from ground-truth actions. The 16 clinical actions (App.~\ref{sec:appendix_actions}; background excluded) are aggregated into 8 macro-categories: Examination (Palpate Wrist, Apical Pulse, Lung Sounds, Temperature, Blood Pressure), Hygiene (Hand Hygiene, Gloves), Screen (Patient History, Vital Signs), Writing, Calculator, Med Bottle, Prep Med, and Apply Med. \textcolor{green!50!black}{Green} edges denote transitions shared by both groups; \textcolor{Red!70!black}{Red} edges are unique to one group. Percentages mean transition probabilities along each arrow.}
\label{fig:process_models}
\end{figure*}

Items related to purely procedural tasks performed in a static, repetitive manner (e.g., Item~1 ``Obtains pertinent data,'' $\rho \approx 0$) show no association, consistent with the expectation that these actions appear visually similar regardless of competency level. Overall, these patterns suggest that vision-based analysis is most informative for expected behaviors tied to procedural diversity and protocol complexity, but is fundamentally limited in capturing the \emph{content} of verbal communication and clinical reasoning.

\subsection{RQ3: Temporal Patterns of Higher vs.\ Lower Performers}
\label{sec:results_rq3}

To understand what distinguishes higher- from lower-performing students, we partition sessions by median competency score and construct process models from ground-truth action sequences (Fig.~\ref{fig:process_models}).

Several structural differences emerge (detailed analysis in App.~\ref{app:process_model_details}). Lower performers show a higher Screen self-loop (48\% vs.\ 41\%), reflecting more time lingering on the bedside monitor, a visually uniform action that inflates MOF. Higher-performing students distribute transitions more evenly across Examination, Writing, and Calculator. The medication pathway also differs: higher performers show a direct Prep Med $\to$ Apply Med transition (46\%), while low performers route through Screen (38\%), suggesting workflow hesitation. Higher-performing students engage in more Examination actions (36 vs.\ 29), which involve diverse movements that are inherently harder to classify, consistent with the observed negative trend between accuracy and competency. Furthermore, the lower-performing students model contains more group-unique (red) transitions, indicating irregular workflow paths, while higher-performing students follow a more protocol-consistent progression. Finally, higher-performing students exhibit a strong Hygiene $\to$ Screen transition (76\%), suggesting more consistent infection-control practices.

To rule out annotation artifacts as the source of this negative trend, we perform a partial association analysis controlling for six potential confounders (annotation coverage, segment count, unique action types, average segment duration, video duration, and total annotations). The MOF--competency association persists across all controls and \emph{strengthens} when controlling for annotation coverage ($\rho$: $-0.439 \to -0.546$, $p = 0.009$); full results are reported in App.~\ref{app:confound}.

\section{Discussion}
\label{sec:discussion}
Our findings raise the question of whether higher frame-level accuracy is always the appropriate optimization target for action recognition in educational settings. In our data, the classifier tends to perform better on sessions with repetitive actions, whereas the more fluid and adaptive workflows associated with higher instructor-rated competency appear harder to recognize. This asymmetry is partly rooted in the prototype-based design: each action class is represented by a single centroid, which favors within-class visual consistency. Students who perform an action similarly across instances produce tighter feature clusters that are easier to match, whereas students who vary their approach across instances produce more dispersed features that weaken prototype fit. One interpretation is that competency, as assessed by clinical educators, includes behavioral diversity and procedural flexibility that current vision models do not fully capture. This view also aligns with broader work suggesting that the extent to which an individual's responses align with a group can relate to learning and memory outcomes; here, the analogous notion of ``fit'' is between a student's action sequence and a prototype-based model constructed from peers through leave-one-out cross-sample prototypes.

This observation suggests a practical two-tier approach to automated assessment: (1)~the predicted action timeline provides a coarse behavioral summary of what the student did, and (2)~the recognition \emph{difficulty} of each session, quantified by mIoU or F1, may serve as a complementary signal of holistic competency. For medication administration, where competency is inherently sequential and correct actions performed in the wrong order can still constitute clinical error, this perspective may help educators identify students whose workflows deviate from the expected procedural pathway even when checklist ratings appear similar. The limited variability in instructor C-CEI ratings further suggests that checklist-based instruments may lack the granularity to distinguish students with clustered overall scores; recognition accuracy may complement such instruments by capturing differences in \emph{how} workflows are executed. More broadly, both rubric-based assessment and vision-based analysis are limited to observable behavior and do not capture the clinical reasoning behind procedural choices. Combining recognition accuracy with post-simulation reflection data, such as structured debriefs or self-assessments, may therefore provide a more complete view of student ability across behavioral and cognitive dimensions. However, with only $N = 22$ sessions, this observation remains exploratory, and accuracy should be viewed as one potential indicator.

\section{Conclusion} \label{sec:conclusion}

We presented a three-stage framework for automated competency assessment from egocentric nursing simulation videos. Our results suggest that recognition accuracy may itself carry a meaningful assessment signal: more competent students produce diverse workflows that are systematically harder to classify. This points to a two-tier assessment perspective, in which predicted action timelines provide a behavioral summary, while recognition difficulty offers a complementary signal of learner competency, clarifying both the promise and the limits of video-based assessment. The primary limitation of this study is the small sample size, reflecting the constraints of privacy-regulated data collection and expert annotation. Future work will investigate protocol-aware competency monitoring using formal methods, where predicted action timelines are checked against procedural specifications to detect missing steps, order violations, and safety-critical deviations. Another important direction is personalized competency trajectory mining, to characterize how different learners develop distinct yet effective behavioral pathways over repeated simulations. More broadly, extending video-based assessment toward multi-level modeling of behavior, communication, and clinical reasoning remains an important open challenge.

\section*{Acknowledgment}
This work was supported in part by the National Science Foundation under grants 2418602 and 2443803. Any opinions, findings, and conclusions or recommendations expressed in this material are those of the authors and do not necessarily reflect the views of the National Science Foundation.

{
    \small
    \bibliographystyle{ieeenat_fullname}
    \bibliography{main}
}

\clearpage
\setcounter{page}{1}
\maketitlesupplementary
\appendix

\section{Simulation Scenario Summary}
\label{app:scenario}

\paragraph{Scenario Context and Setting}
The simulation scenario and debrief were created by a nursing teaching instructor and have been used in nursing school classroom settings. 
The simulation scenario takes place in a high-fidelity pediatric emergency room bay. A standardized pediatric manikin representing a toddler and a faculty facilitator acting as the patient's caregiver are present at the bedside. The scenario is designed to evaluate pediatric assessment, weight-based medication administration, and caregiver communication competencies.

\paragraph{Anonymized Patient Profile}
The simulated patient is a 16-month-old toddler (9.6~kg, 76~cm) presenting with a primary diagnosis of croup (laryngotracheobronchitis). The caregiver reports a 3-day history of upper respiratory infection symptoms, with sudden overnight onset of a barking cough, hoarse voice, and inspiratory stridor. On arrival, the patient is placed on continuous pulse oximetry, heart rate, and respiratory rate monitoring, and maintained on humidified oxygen at 1~LPM via pediatric face mask.

\paragraph{Simulation Learning Objectives}
The scenario targets four core nursing competencies: (1) performing a focused pediatric assessment while maintaining age-appropriate patient safety; (2) recognizing pediatric fever and calculating accurate weight-based dosages for oral antipyretic medications; (3) preparing and administering pediatric oral suspensions safely; and (4) providing clear, developmentally appropriate education to caregivers regarding at-home medication administration.

\paragraph{Scenario Progression and Key Interventions}
The simulation unfolds across three phases. In \textbf{Phase 1}, the student initiates care by performing hand hygiene, verifying patient identification using two identifiers, and introducing themselves to the caregiver. Initial vitals reflect tachypnea and tachycardia consistent with the patient's respiratory distress. In \textbf{Phase 2}, the student performs a focused respiratory assessment, noting mild expiratory wheezing and intermittent barking cough. A bedside temperature check reveals a fever of $102.6^\circ\text{F}$, prompting the student to review physician orders and perform a weight-based medication calculation for oral acetaminophen suspension ($160~\text{mg}/5~\text{mL}$):
\begin{align}
15~\text{mg/kg} \times 9.6~\text{kg} &= 144~\text{mg}, \notag \\
144~\text{mg} \times \frac{5~\text{mL}}{160~\text{mg}} &= 4.5~\text{mL}.
\end{align}
In \textbf{Phase 3}, after preparing the medication in an amber oral dosing syringe, the student engages in targeted caregiver education. Key instructional points include advising against using household spoons for measuring, demonstrating correct syringe administration technique to prevent choking, and establishing safe guidelines for dosing frequency at home.

\section{Instructor Competency Rubric}
\label{sec:appendix_rubric}

The instructor rubric is an adapted version of the Creighton Competency Evaluation Instrument (C-CEI)~\cite{todd2008ccei}, which maps 23 expected behaviors to broader concepts of competency (e.g., clinical judgment, patient safety, communication). Each item is rated on a 1--5 scale (Poor to Exceptional). Because our study uses egocentric video without audio, only the \textbf{11 video-observable items} (highlighted in Tab.~\ref{tab:rubric_full}) contribute to each student's competency percentage. The remaining 12 items require verbal or cognitive assessment not accessible from visual data alone.

\begin{table}[ht]
\centering
\caption{Full 23-item C-CEI rubric. Each item specifies an expected behavior; highlighted rows ($\checkmark$) are the 11 video-observable items used for competency scoring.}
\label{tab:rubric_full}
\resizebox{\columnwidth}{!}{%
\begin{tabular}{clc}
\toprule
\# & Item Description & Video? \\
\midrule
\rowcolor{green!8}
1  & Obtains pertinent data & $\checkmark$ \\
\rowcolor{green!8}
2  & Performs follow-up assessments as needed & $\checkmark$ \\
3  & Assesses the environment & $\times$ \\
4  & Communicates effectively with team & $\times$ \\
5  & Communicates effectively with patient & $\times$ \\
\rowcolor{green!8}
6  & Documents clearly, concisely, and accurately & $\checkmark$ \\
7  & Responds to abnormal findings appropriately & $\times$ \\
8  & Promotes professionalism & $\times$ \\
\rowcolor{green!8}
9  & Interprets vital signs & $\checkmark$ \\
10 & Interprets laboratory results & $\times$ \\
11 & Interprets subjective/objective data & $\times$ \\
\rowcolor{green!8}
12 & Prioritizes appropriately & $\checkmark$ \\
\rowcolor{green!8}
13 & Performs evidence-based interventions & $\checkmark$ \\
14 & Provides evidence-based rationale for interventions & $\times$ \\
15 & Evaluates evidence-based interventions and outcomes & $\times$ \\
16 & Reflects on clinical experience & $\times$ \\
17 & Delegates appropriately & $\times$ \\
\rowcolor{green!8}
18 & Uses patient identifiers & $\checkmark$ \\
\rowcolor{green!8}
19 & Utilizes standardized practices and precautions & $\checkmark$ \\
\rowcolor{green!8}
20 & Administers medications safely & $\checkmark$ \\
\rowcolor{green!8}
21 & Manages technology and equipment & $\checkmark$ \\
\rowcolor{green!8}
22 & Performs procedures correctly & $\checkmark$ \\
23 & Reflects on potential hazards and errors & $\times$ \\
\bottomrule
\end{tabular}%
}
\end{table}

\noindent\textbf{Rationale for the video-observable subset.}
Items requiring verbal content (e.g., ``Communicates effectively with team,'' ``Provides evidence-based rationale'') or internal cognitive processes (e.g., ``Reflects on clinical experience'') cannot be assessed from silent egocentric video. The 11 retained items correspond to physical actions and procedural behaviors that produce visible evidence in the video stream: checking wristbands, performing hand hygiene, documenting on screens, measuring vital signs, administering medications, and following safety protocols. This principled subset ensures that the competency score reflects only expected behaviors that our vision-based system could plausibly detect. Note that in the per-item association analysis (Tab.~\ref{tab:per_item}), we report associations for all 23 items to explore whether vision-based features carry any indirect signal for non-observable behaviors.

\section{Action Annotation Rubric}
\label{sec:appendix_actions}

The following is from our annotation codebook, used by trained coders to produce ground-truth action annotations, and is inspired by \cite{schroers2025tool}. Actions represent discrete, observable physical behaviors; verbal introductions are captured separately by the Communication layer.

\smallskip
\noindent\textbf{General coding rules.}
\begin{enumerate}[leftmargin=*,nosep]
    \item Code only what is directly observable; do not infer intent.
    \item When in doubt, leave the segment unlabeled.
    \item Annotations must not overlap within the Action layer.
    \item Start when the action begins (first observable movement); end when it concludes (hands leave the object, body repositions away).
    \item Brief interruptions ($<$2\,s): code as one continuous segment.
\end{enumerate}

\smallskip
\noindent\textbf{Action definitions.}
Tab.~\ref{tab:action_vocab} lists the $K{=}16$ fine-grained clinical action classes. Frames that do not correspond to any of these classes (e.g., walking, adjusting equipment, idle periods between clinical actions) are left unannotated and treated as the background class $a_\varnothing$, yielding $K{+}1{=}17$ labels in total for recognition.

\begin{table}[ht]
\centering
\caption{The $K{=}16$ clinical action classes used for temporal annotation and few-shot recognition, with brief operational definitions. An additional background class $a_\varnothing$ (not shown) captures all non-clinical frames, yielding 17 labels total.}
\label{tab:action_vocab}
\resizebox{\columnwidth}{!}{%
\begin{tabular}{cll}
\toprule
ID & Action Class & Definition \\
\midrule
1  & Perform Hand Hygiene & Uses hand sanitizer or washes hands at sink \\
2  & Put on Gloves & Retrieves and dons disposable gloves \\
3  & Check Patient Wristband & Visually inspects or scans patient wristband \\
4  & Check Patient History Screen & Reads electronic health record on screen \\
5  & Examine Med Bottle & Picks up and reads medication label \\
6  & Review Vital Signs Screen & Reads the vital signs monitor (HR, BP, SpO$_2$) \\
7  & Assess Vital Signs (Palpate Wrist) & Manually palpates radial pulse \\
8  & Auscultate Lung Sounds & Places stethoscope on chest/back for breath sounds \\
9  & Measure Apical Pulse & Places stethoscope at heart apex, held $\geq$15\,s \\
10 & Measure Temperature & Uses thermometer (oral, tympanic, temporal) \\
11 & Measure Blood Pressure & Initiates BP reading via monitor or manual cuff \\
12 & Writing & Pen-to-paper: notes, calculations, forms \\
13 & Use Calculator & Computes dosage on physical or phone calculator \\
14 & Check Phone & Interacts with phone for non-calculator purposes \\
15 & Prepare Medication & Draws syringe, crushes tablet, mixes solution \\
16 & Apply Medication to Patient & Administers medication: oral, IV, injection, topical \\
\bottomrule
\end{tabular}%
}
\end{table}

\smallskip
\noindent\textbf{Disambiguation guidelines.}
Several action pairs are visually similar and require explicit decision rules:

\smallskip
\noindent\textit{Lung Sounds (\#8) vs.\ Apical Pulse (\#9):}
Stethoscope on the back or moved across multiple chest positions is coded as \#8.
Stethoscope held at the left chest apex in one position for $\geq$15\,s is coded as \#9.
If placement is unclear, default to \#8 and flag for review.

\smallskip
\noindent\textit{Calculator (\#13) vs.\ Phone (\#14):}
Tapping numbers on a calculator app or physical calculator is \#13.
Scrolling, reading, or swiping on a phone (non-calculator) is \#14.

\smallskip
\noindent\textit{Patient History Screen (\#4) vs.\ Vital Signs Screen (\#6):}
If the screen shows waveforms or real-time numeric readings, code as \#6.
If it shows text-based records, history, or medication orders, code as \#4.
Pressing a button on the vitals monitor to initiate a BP measurement is coded as \#11.

\section{Process Model Details}
\label{app:process_model_details}

The five key structural differences between higher- and lower-performing students process models (Fig.~\ref{fig:process_models}) are elaborated below.

\noindent\textbf{Screen self-loop.} Lower-performing students exhibit a higher self-loop on the Screen action (48\% vs.\ 41\%), spending proportionally more time returning to the bedside monitor without transitioning to other clinical actions. Higher performing students distribute transitions away from Screen more evenly across Examination, Writing, and Calculator, reflecting a more fluid workflow. Screen actions are visually static and uniform, making them easy for the classifier and inflating MOF for the lower group.

\noindent\textbf{Medication pathway.} Higher-performing students show a strong direct Prep Med $\to$ Apply Med transition (46\%), indicating a coherent prepare-then-administer sequence. Lower-performing students lack this link; instead, Prep Med routes back to Screen (38

\noindent\textbf{Examination frequency.} Higher performers engage in more Examination actions (36 vs.\ 29), while lower-performing students produce more Writing and Screen actions (42 and 79 vs.\ 37 and 74). Physical examination (lung sounds, blood pressure, palpation) involves diverse movements inherently harder to classify, consistent with the observed negative trend between accuracy and competency.

\noindent\textbf{Transition irregularity.} The lower-performing students model contains more group-unique (red) transitions, indicating irregular workflow paths. Higher performers follow a more protocol-consistent progression with fewer idiosyncratic transitions.

\noindent\textbf{Hygiene compliance.} Hygiene actions connect to Screen with 76\% probability in higher performing students, suggesting consistent hand hygiene before engaging with the patient monitor. This transition is less prominent in lower-performing students, pointing to less consistent infection control practices.

These process model comparisons offer actionable insight for clinical educators: the transition graphs visualize where each student's workflow diverges from the expected clinical pathway, enabling targeted remediation of specific procedural gaps.

\section{Annotation Confound Analysis}
\label{app:confound}

A potential concern is that annotation artifacts drive the negative trend between classification accuracy and competency, since higher-competency sessions have lower annotation coverage (40\% vs.\ 50\%). We perform partial association analysis, controlling for six potential confounds (Tab.~\ref{tab:partial}). If any drove the observed pattern, controlling for it would weaken or eliminate the effect.

\begin{table}[ht]
\centering
\caption{Robustness analysis. \emph{Partial $\rho$}: Spearman association between MOF and competency after controlling for each variable. \emph{Var $\leftrightarrow$ MOF}: bivariate association between each variable and MOF. The MOF--competency association persists across all controls and \emph{strengthens} when controlling for annotation coverage (bolded). No control variable independently predicts MOF (all $p > 0.46$).}
\label{tab:partial}
\scriptsize
\begin{tabular}{lcccc}
\toprule
Control Variable & Partial $\rho$ & $p$ & Var $\leftrightarrow$ MOF $\rho$ & $p$ \\
\midrule
None (baseline)            & $-0.439$ & $0.041$ & --- & --- \\
\textbf{Annotation coverage} & $\mathbf{-0.546}$ & $\mathbf{0.009}$ & $-0.032$ & $0.887$ \\
\# GT action segments      & $-0.427$ & $0.047$ & $+0.115$ & $0.611$ \\
\# Unique GT action types  & $-0.438$ & $0.041$ & $+0.027$ & $0.907$ \\
Avg segment duration       & $-0.437$ & $0.042$ & $-0.165$ & $0.462$ \\
Video duration             & $-0.454$ & $0.034$ & $+0.074$ & $0.744$ \\
\# All annotations         & $-0.427$ & $0.047$ & $+0.115$ & $0.611$ \\
\bottomrule
\end{tabular}%
\end{table}

The pattern persists across all controls. When controlling for annotation coverage, the effect \emph{strengthens} ($\rho$: $-0.439 \to -0.546$), and no control variable independently predicts MOF (all $p > 0.46$), confirming that the negative trend reflects workflow complexity rather than annotation density.

\section{Inter-Rater Reliability}
\label{app:irr}

A second rater independently annotated 3 stratified videos (low / median / high competency) to assess annotation reliability. Agreement was measured using frame-level Cohen's $\kappa$ at 1~Hz resolution. To avoid inflation from unannotated frames, $\kappa$ was computed only over frames where at least one rater placed a label.

Mean $\kappa = 0.708 \pm 0.199$ (substantial agreement; \cite{landis1977measurement}). As a secondary metric, mean per-class IoU $= 0.697 \pm 0.143$, and both raters identified identical action type sets in all 3 videos (Jaccard $= 1.0$). Disagreements were predominantly in segment boundary placement, particularly action endpoints (mean $|\Delta| = 3.4$~s), rather than action identification or ordering. This pattern indicates that raters agree on \emph{which} actions occur and \emph{in what order}, with variability confined to the precise temporal boundaries, consistent with the known difficulty of endpoint annotation in temporal action segmentation~\cite{ding2024temporal}.

\end{document}